\documentclass[10pt, a4paper]{article}

\usepackage[final]{lrec2026} % this is the new style
\usepackage{booktabs}
\usepackage{amsmath}
\usepackage{cleveref}
\usepackage{pgfplots}

\usepackage{pgfplotstable}
\usepackage{tikz}
\usepgfplotslibrary{colormaps}
\usepackage[utf8]{inputenc}
\usepackage{graphicx}

\usepackage{subcaption}
\usepackage{booktabs}
\usepackage{adjustbox}
\usepackage{enumitem}
\pgfplotsset{compat=1.18}
\usepackage[most]{tcolorbox}

\title{Language Ideologies in a Multilingual Society: \\ An LLM-based Analysis of Luxembourgish News Comments}
\name{Emilia Milano\textsuperscript{1}, Alistair Plum\textsuperscript{1}, Yves Scherrer\textsuperscript{2}, Christoph Purschke\textsuperscript{1}} 

\address{\textsuperscript{1}University of Luxembourg, Esch-sur-Alzette, Luxembourg, \\
         \textsuperscript{2}University of Oslo, Oslo, Norway \\
         \{emilia.milano, alistair.plum, christoph.purschke\}@uni.lu\\
         yves.scherrer@ifi.uio.no
         }

\abstract{
Detecting language ideologies is a valuable yet complex task for understanding how identities are constructed through discourse. In Luxembourg’s multicultural and multilingual society, language ideologies reflect more than simple preferences: they carry deep cultural and social meanings, shaping identities and social belonging. Following recent developments in applying Natural Language Processing tools to linguistics and social science, this paper explores the potential of large language models to assist in the detection of language ideologies.
We manually annotate a corpus of user comments in Luxembourgish with predefined ideological categories and then evaluate the performance of large language models under varying prompt conditions to assess their ability to replicate these human annotations. Since Luxembourgish is a small language and poorly represented in the LLMs' training data, we also investigate whether machine-translating the data to high-resource languages increases performance on the ideology detection task. 
Our findings suggest that, while LLMs are not yet fully optimized for a multi-class ideological annotation task, they are practical tools to identify language ideological content. %163 words
 \\ \newline }

\begin{document}

\maketitleabstract
\section{Introduction}

Automatic detection of language ideologies is a promising but challenging task. Language ideologies, or shared beliefs about language, play a central role in reinforcing and establishing identities and power imbalances \cite{blommaert_language_2010}. Although culturally tied to the speakers' community, similar language ideologies have been found in different contexts. Often, language ideologies are linked to matters of social belonging and discrimination \cite{grondelaers2011, lippi2012english}. 
This is also found in our case study, Luxembourg. The country is characterized by a complex and dynamic societal multilingualism, in which the choice of language in interaction carries rich social meaning aside from practical requirements or personal preference \cite{hornerweber2008}. This is evidenced in public discourses on the state of multilingualism and the role of Luxembourgish, where language choices and ideological stances are often closely linked. For example, the phrase \textit{En français, s'il vous plaît} (`In French, please'), used by (French-speaking) service personal as a reaction to requests in Luxembourgish, has become a prominent figure in discourse \cite{purschke2025}: It is interpreted (and quoted) by speakers of Luxembourgish as a sign of the unwillingness to integrate in society, linguistically and socially, and linked to questions of demographic development and language vitality \cite{purschke2023diskurs}.

%where the language use in formal or informal communications does not only become a practical choice, but mainly an identity one. Using the national language (Luxembourgish), instead of the other two official ones, states the identity of the speaker as someone belonging in the local society. Along the same lines, who is not able to speak Luxembourgish is labeled as an outsider.
%the expression is often used for practical reasons by people not speaking the national language \cite{purschke2025}. For speakers of Luxembourgish, this became a shibboleth to distance themselves from non-Luxembourgish speakers, complain about the lack of adaptation by part of the out-group to the local society, and about the strong presence of French in the country.

Hence, being able to automatically and systematically detect language ideologies in large amounts of text would be beneficial for obtaining a deeper understanding of cultural dynamics and language-related discourse at scale. 

Against this backdrop, the increasing use of Large Language Models (LLMs) for linguistic \cite{manning_emergent_2020, wu2025large} and social science research \cite{ma_potential_2024} presents a promising opportunity to systematically investigate the automatic detection and classification of language ideologies.

For this study, we base ourselves on a corpus of written Luxembourgish comments provided by the national broadcaster RTL\footnote{Our agreement with the data provider prevents us from publicly releasing the corpus. However, we can provid upon request the comment IDs so that the corpus can be recreated.}\cite{ranasinghe2023publish}. We annotate the comments at two levels: (1) whether they are related to language ideology or not, and (2) which different types of language ideologies they contain (see Section \ref{sec:ideology_categories}). We assess a range of open-weights and closed-weights LLMs on their ability to predict these annotations, testing several prompts in both zero-shot and few-shot settings.

Since Luxembourgish is a small language, we presume that it is comparatively poorly represented in the LLMs' training data and that this results in low classification performance. We therefore follow the \textit{translate-test} paradigm and investigate whether machine translating the Luxembourgish comments into a higher-resourced language (specifically, English, French or German) increases classification accuracy.

More specifically, our research questions are:
\begin{enumerate}
\item Is language ideology classification a suitable task for the current generation of LLMs? 
\item Do the languages of the data and the quality of their translations have an impact on LLM performance on language ideology classification?
\item Can LLMs contribute to inform sociolinguistic research without requiring high technical expertise or task-specific model adaptation?
\end{enumerate}

To explore the feasibility and challenges of the task, we provide LLMs with language ideology categories, descriptions and examples, and instructions to annotate accordingly. Our contributions offer an evaluation of LLMs for language ideologies identification, assessing both the impact of the source language and translations on the same task. In doing so, we provide an overview of how LLMs work with Luxembourgish on cultural tasks, compared to more established languages in NLP. We find that fine-grained language ideology detection is a complex task for LLMs, and that translations do not lead to substantial performance improvement. Integrating NLP tools into sociolinguistic research, we acknowledge limitations and possible improvements of the task setup.

\section{Related work}

In this section, we discuss work related to Luxembourgish NLP (Section \ref{sec:rel1}), as well as language ideologies (Section \ref{sec:ideology_categories}). After that, we discuss work on ideologies and LLMs (Section \ref{sec:rel3}), as well as cross-lingual transfer in LLMs (Section \ref{sec:rel4}).

\subsection{Luxembourgish}\label{sec:rel1} 

Luxembourgish is a West Germanic language with about 400.000 speakers \cite{Gilles2019}. It is one of three official languages in Luxembourg in addition to German and French, but it was only recognized as such in 1984. 
Historically, Luxembourgish mainly developed as an oral language, acquiring more written domains only recently with the rise of digital and social media. As a multilingual country with an extremely dynamic language regime, Luxembourg is a particularly interesting case for sociolinguistic studies \cite{purschke2023sociolinguistics}. However, in NLP, Luxembourgish should be viewed as a small language, although various resources have been made available in recent years \cite{plum2024text, lutgen2024neural, philippy-etal-2024-forget, plum2024luxbank}.

\subsection{Language ideologies}\label{sec:ideology_categories}

We follow \citet{blume2003}'s description of ideologies and understand them as descriptions of collective, depersonalized normative interpretations used by individuals and groups to justify and evaluate their own and others' actions. As such, ideologies towards and about language \cite{blommaert_language_2010} play a central role in creating and ordering social meaning \cite{rhodes2023language}, in shaping language policies and perception, in structuring social relations and societal belonging \cite{stanlaw_language_2020}, and in contributing to individual and collective identity constructions \cite{volkel_5_2022}. 

From a methodological perspective, (language) ideologies have also been studied with quantitative tools and methods. \citet{baker2015picking} compares qualitative and corpus-based approaches to the analysis of newspaper articles on masculinity, showing that the two approaches can complement each other, and \citet{vessey2017corpus} investigates language ideologies in Canada through corpus linguistics approaches. %A corpus linguistic approach has been used to investigate language ideologies \cite{vessey2017corpus}. 
Combining different approaches to corpus study and language ideologies, \citet{purschke2025} analyzes language ideologies on the RTL.lu corpus as part of public discourse formation. 

Building on this foundation, we search for the following five language-ideological motifs about Luxembourgish and multilingualism in the country in the RTL.lu comments. %The ideological motifs were empirically detected on a small subset (10 comments) of the corpus in analysis.%\footnote{We will provide examples of the five motifs in the appendix of the camera-ready version}: 

\begin{description}
\item[Identity:] statements about national identity established by a common language, culture, and history, or personal preference towards the national language;\\
Example ID 636: \textit{`1. For us Luxembourgers (250,000 people), Luxembourgish is our first language.'}\footnote{Original texts are provided in the appendix.}
\item[Vitality:] statements about the state of a language use, e.g., language endangerment and decay, reasons for these developments and/or means to prevent them;\\
Example ID 3631: \textit{`Good text, but I'm afraid our language would die out.'}
\item[Belonging:] statements about requirements for speakers to integrate in the local society;\\
Example ID 410165: \textit{`it worked: all residents of Luxembourg have linguistically integrated themselves.'}
\item[Responsibility:] statements delegating to an actor or a group of actors responsibility for the linguistic situation in the country (e.g. language decay or support);\\
Example ID 666: \textit{`And if our cross-border commuters don't speak Luxembourgish, it's our own fault.'}
\item[Recognition:] statements about language policies, language status, and how people perceive the languages spoken in the country.\\
Example ID 636: \textit{`2. Luxembourgish is FIRST OF ALL a spoken language.'}

\end{description}

\subsection{Ideologies and LLMs}\label{sec:rel3}

There is abundant work showing how human stereotypes, biases, and ideologies influence the performance of (large) language models \cite{lin2025implicit}. They can be introduced in the (pre-)training data \cite{DBLP:journals/llc/HovyP21}, optimization stages \cite{dahlgren2025helpful}, and data exposure level \cite{chen_how_nodate}, they can change according to the tested model \cite{lin_investigating_2024} and language \cite{buyl_large_2025}, and can be shaped by language ideologies. For instance, \citet{gururangan_whose_2022} investigate the role of language ideologies in selecting text data and the subsequent inequalities. LLMs have also been shown to perpetuate in-group solidarity and out-group hostility \cite{hu_generative_2024} -- where in-group refers to the community speakers socially and emotionally identify with, and out-group refers to communities perceived as outsiders.
The language material generated by LLMs has been shown to reproduce social bias \cite{dhamala_bold_2021} as well as language ideologies, both for the standard language ideology in the US \cite{smith_standard_2025} and for gendered language ideology in English \cite{watson-etal-2025-language} and English dialects \cite{duncan2024does}. 

\subsection{Cross-lingual transfer in LLMs}\label{sec:rel4}
Luxembourgish can be considered a small language in a multilingual context characterized by higher resourced languages. Almost all Luxembourgers are fluent in the other official languages, French or German \cite{fehlen2023diversite}, and despite Luxembourg's membership in the European Union, Luxembourgish is not an official EU language. Additionally, Luxembourgish is not the language of education, and its domain expansion and societal uptake in writing are relatively recent \cite{gilles2015}. For all these reasons, the amount of textual data to train large-scale NLP systems is limited. In such scenarios, it is common to resort to cross-lingual transfer techniques, ensuring that a model's high performance on a specific task in a high-resource language can be transferred to a low-resource language. A simple cross-lingual transfer technique that can be easily applied in the context of generative LLMs is called \textit{translate-test} \cite{ponti2021modellinglatenttranslationscrosslingual,artetxe-etal-2023-revisiting}: the test data (in our case, the user comments written in Luxembourgish) are machine-translated to a high-resource language (in our case, English, French or German) and then fed to the model. While this setup is prone to error propagation (the translations might contain errors), it is assumed that the performance gains obtained through the model's better ``understanding'' of the translated data outweigh the errors introduced in the translation step.

\section{Data}

We frame ideology detection as a multi-class sentence annotation task over Luxembourgish user comments from the RTL.lu corpus. We operate at sentence level to reduce topic drift within long comments. 
Our corpus contains 300 comments with 1524 total sentences. Comments were split into sentences using GPT, and then manually corrected where the sentence splitting did not work correctly. The main reason for using GPT for this task is that, for social media data with irregular sentence constructions and punctuation patterns (e.g., including Emoji as sentence boundaries) other sentence splitters usually struggle and GPT, compared to these, includes semantic context parsing for better accuracy.
We retained the punctuation by each user and did not carry out any other pre-processing steps such as lexical normalization, as these could blur ideology markers. In a setting without a fully implemented standard, the use of incorrect or correct orthography could be an expression of specific language ideologies. 

The corpus is extracted from a larger set of user comments (1,422,759 comments at the time of the study\footnote{For this study, we used the 2025-04 build of the RTL corpus.}). It contains two subsets, one subset of 150 comments on language use in the country (from now on, we will refer to this subset as \textit{Lang}), and one subset of 150 comments not related to this topic (referred to as \textit{NotLang}). 
For the Lang subset, the selection of comments was made through keyword matching, using a total of 19 Luxembourgish variants of \textit{Luxembourgish, our language, mother tongue}. Of the comments retrieved (16,015), 150 are then randomly selected and manually annotated.

For the NotLang subset, 75 comments were selected following the same list of keywords as for the Lang subset, and were carefully read to make sure that they refer to topics other than language use in the country. The remaining 75 comments are randomly selected from the complete set of comments (1,422,759), and do not contain language-related keywords. By diversifying the selection criteria across subsets, we aim to prevent the models from overfitting to surface-level lexical cues and to assess their performance in identifying ideologically charged discourse about language.

The Lang subset is annotated by two expert annotators (the first and last authors of this study) with the five language ideology categories presented in Section~\ref{sec:ideology_categories}: \textsc{identity}, \textsc{vitality}, \textsc{belonging}, \textsc{responsibility} and \textsc{recognition}. We use a sixth label \textsc{none} for sentences that cannot be assigned one of these specific labels. The observed agreement between the two annotators is 0.78 and Cohen's Kappa is 0.66.
Disagreement cases are discussed by annotators and are labeled in common agreement. 
The category counts across the dataset are shown in \autoref{tab:annotation_stats}. %, and confusion matrix for the manual annotation of the Lang subset in \autoref{fig:annotators} . 

\begin{table}[ht]
\centering
\footnotesize
\begin{tabular}{lcc}
\toprule
\textbf{Category} & \textbf{Lang} & \textbf{NotLang} \\
\midrule
Identity       & 49   & 0   \\
Vitality       & 72   & 0   \\
Belonging      & 63   & 0   \\
Responsibility & 121  & 0   \\
Recognition    & 125  & 0   \\
None            & 273 & 821 \\
\midrule
\textbf{Total} & \textbf{703} & \textbf{821} \\
\bottomrule
\end{tabular}
\caption{\label{tab:annotation_stats}Sentence based gold standard annotation}
\end{table}

%\begin{figure} [ht]
%    \centering
%    \includegraphics[width=7cm]{confmatrix_ANNOTATORS.png}
%    \caption{Confusion matrix globally normalized of the manual annotations of Lang subset}
%    \label{fig:annotators}
%\end{figure}

\begin{figure*}
\centering
\footnotesize
\begin{tabularx}{\textwidth}{|X|}
\hline
\textbf{Prompt Text} \\ \hline
Detect one or more of the following categories in the sentences of the comment below, after the instructions and schemas. The categories are: Identity, Vitality, Belonging, Responsibility, Recognition. These categories concern language ideologies and only these exact categories can be used for the annotation. No overlap between categories is allowed, each sentence can have 0 or 1 category. If you cannot identify one of the category assign N/A. Do not take any quoted parts (i.e. text in quotation marks) into account for the labelling. For each sentence, explain why you have assigned that label. \\

Here are the explanations of the five categories: [...] \\ \hline

\textbf{Recognition:} The language related ideology ‘recognition’ groups opinion on Luxembourgish language and other languages spoken in the country. In this category, different ways of acknowledging languages are involved:  [...]\\
\textbf{Examples:}  [...]\\
\textbf{Examples where this label is not applicable:} 
[...] \\ \hline

\end{tabularx}
\caption{\label{tablex} Prompt 4 comprehensive of instructions, definitions, and examples}
\end{figure*}

% \begin{table*}
% \centering
% %\resizebox{\linewidth}{!}{
% \resizebox{0.9\textwidth}{!}{ 
% %\renewcommand{\arraystretch}{1.2}
% \begin{tabularx}{\textwidth}{|X|}
% \hline
% \textbf{Prompt Text} \\ \hline
% Detect one or more of the following categories in the sentences of the comment below, after the instructions and schemas. The categories are: Identity, Vitality, Belonging, Responsibility, Recognition. These categories concern language ideologies and only these exact categories can be used for the annotation. No overlap between categories is allowed, each sentence can have 0 or 1 category. If you cannot identify one of the category assign N/A. Do not take any quoted parts (i.e. text in quotation marks) into account for the labelling. For each sentence, explain why you have assigned that label. \\

% Here are the explanations of the five categories: [...] \\ \hline

% \textbf{Recognition:} The language related ideology ‘recognition’ groups opinion on Luxembourgish language and other languages spoken in the country. In this category, different ways of acknowledging languages are involved:  [...]\\
% \textbf{Examples:}  [...]\\
% \textbf{Examples where this label is not applicable:} 
% [...] \\ \hline

% \end{tabularx}}
% \caption{\label{tablex} Prompt 4 comprehensive of instructions, definitions, and examples}
% \end{table*}

\section{Prompt Engineering} \label{sec:prompts}

In this first experiment, we focus on evaluating different prompts in both zero- and few-shot scenarios.

\paragraph{Prompts} We design four prompt configurations that vary the amount of information that is passed to the model.
\begin{itemize}
    \item \textbf{Prompt 1} lists the labels, instructs the LLM to annotate each sentence with one of the five ideology labels or \textsc{none} and to provide a justification for the annotation. Text in quotation marks should not be annotated.
    \item \textbf{Prompt 2} adds three positive and three negative examples per category, and six additional negative examples to prompt 1. The examples are chosen because they are clear illustrations for the categories and are consistent throughout the experiment.
    \item \textbf{Prompt 3} adds detailed descriptions per category to prompt 1.
    \item \textbf{Prompt 4} adds both the examples of prompt 2 and the detailed descriptions of prompt 3 to prompt 1.
\end{itemize}

The structure of prompt 4, the most complex one, is illustrated in \autoref{tablex}. All prompts are written in English, and all comments remain in Luxembourgish.
All prompts instruct models to return a single JSON object with fixed keys and value constraints, see \autoref{fig:annotation-json-example} below. %We reject free-form text and re-query on schema violations.%, see \autoref{fig:annotation-json-example} below.

\begin{figure}[h]
\centering
\fbox{
\begin{minipage}{0.9\linewidth}
\footnotesize
\setlength{\parskip}{2pt}
\texttt{\\
\textcolor{blue!60!black}{\textquotedbl sentence\_id\textquotedbl}: \textcolor{brown!70!black}{\textquotedbl <string>\textquotedbl},\\
\textcolor{blue!60!black}{\textquotedbl text\textquotedbl}: \textcolor{brown!70!black}{\textquotedbl <original sentence>\textquotedbl},\\
\textcolor{blue!60!black}{\textquotedbl labels\textquotedbl}: [%
\textcolor{brown!70!black}{\textquotedbl Identity\textquotedbl} \textcolor{gray}{|} 
\textcolor{brown!70!black}{\textquotedbl Vitality\textquotedbl} \textcolor{gray}{|} 
\textcolor{brown!70!black}{\textquotedbl Belonging\textquotedbl} \textcolor{gray}{|} \textcolor{brown!70!black}{\textquotedbl Responsibility\textquotedbl} \textcolor{gray}{|} \textcolor{brown!70!black} {\textquotedbl Recognition\textquotedbl} \textcolor{gray}{|} \textcolor{brown!70!black}{\textquotedbl None\textquotedbl}],\\
\textcolor{blue!60!black}{\textquotedbl rationale\textquotedbl}: \textcolor{brown!70!black}{\textquotedbl <justification>\textquotedbl}\\
}

\end{minipage}
}
\caption{JSON representation of the required schema.}
\label{fig:annotation-json-example}
\end{figure}

%\begin{verbatim}
%{
%"sentence_id": "<string>",
%"text": "<original sentence>",
%"labels": ["Identity" | "Vitality" | 
%    "Belonging" | "Responsibility" | 
%    "Recognition" | "None" ],
%"rationale": "<justification>"
%}
%\end{verbatim}

\paragraph{Experimental Setup}  \label{paragraph : setup} We treat the task as multi-class annotation with sentence level ground truth derived from the human annotation. We report weighted F1, macro F1, and per label F1. We implement quality control to improve robustness without over-correcting model behavior. We validate every response against the JSON schema and re-issue malformed outputs up to 5 retries per sentence. 

We test three models of the GPT family: GPT-4o \cite{openai2023gpt4}, GPT-4o-mini \cite{openai2024gpt4omini}, o3 \cite{openai2025gpto3}. We standardize decoding to reduce variance induced by sampling. Temperature is set to 0.2, maximum tokens to 3000, however, we keep defaults for safety-related flags that affect refusal behavior. We batch requests with one comment (5 sentences on average) per call. Also, we enforce per thread rate limits of rps/5 to avoid provider throttling.  
\paragraph{Results} We observe minimal differences between GPT-4o, GPT-4o-mini, and o3 under prompt 4. This prompt also consistently outperforms other configurations across models, as F1 weighted score improves up to 
around 0.3 from prompt 1 to prompt 4 -- see \autoref{table F1/prompts}.
Therefore, we adopt prompt 4 as the final prompting template for all further experiments.  

\begin{table}[ht]
\centering
\resizebox{\linewidth}{!}{
\begin{tabular}{lcccccc}
\toprule
\textbf{Prompt}& \multicolumn{2}{c}{\textbf{o3}} & \multicolumn{2}{c}{\textbf{GPT-4o}} & \multicolumn{2}{c}{\textbf{GPT-4o-mini}} \\
& F1w & F1m & F1w & F1m & F1w & F1m \\
\midrule
Prompt 1 & 0.734 & 0.431 & 0.491 & 0.248 & 0.563 & 0.280 \\
Prompt 2 & 0.787 & 0.510 & 0.652 & 0.397 &  0.459 & 0.389 \\
Prompt 3 & 0.806 & 0.567 & 0.765 & 0.427 & 0.468 & 0.391 \\
Prompt 4 & \textbf{0.806} & 0.559 & \textbf{0.783} & 0.441 & \textbf{0.734} & 0.427\\
\bottomrule
\end{tabular}
}
\caption{\label{table F1/prompts}Results per model and prompting conditions (Weighted F1 and Macro F1)}
\end{table}

\section{Evaluating LLMs and cross-lingual transfer settings} 

In this main experiment, we evaluate several widely used generative LLMs on the ideology annotation task with Prompt 4. We also vary the language of the comments, comparing annotation performance on the original Luxembourgish comments with their translations to English, French and German.

\paragraph{Machine translation} To answer research question 2, we translate the original material into German, French and English via the Google Translate API. The automatic translations receive targeted manual checks\footnote{Automatic translation quality for Luxembourgish can be uneven, therefore, subtle stance cues may shift.} that correct mistranslated country and language mentions. Additionally, 700 sentences evenly split between Lang and NotLang are manually reviewed to improve translation quality; correcting mistakes due to orthographic variation and lack of capitalization, and adapting ideology bearing content to the target languages.
The distribution of categories within the two translation sets is described in \autoref{translated data}.
We run the same prompting and inference pipeline on each language set. Therefore, any observed changes can be attributed to language rather than prompt drift. 

\begin{table}[htbp]
\centering
\footnotesize
\begin{tabular}{lcc}
\toprule
\textbf{Labels}  & \textbf{Post-Edited} & \textbf{Automatic} \\
\midrule
Identity & 18 & 31 \\
Vitality & 22 & 50 \\
Belonging & 16 & 47 \\
Responsibility & 64 & 57 \\
Recognition & 56 & 69 \\
None & 525 & 569 \\
\bottomrule
\end{tabular}
\caption{\label{translated data}Translation sets}
\end{table}

\paragraph{Models}  We evaluate widely used frontier and open generative LLMs for applied annotation workflows. The frontier group includes GPT-5 \cite{openai2025gpt5}, while the open group includes Aya Expanse-32B (Aya; \citeauthor{dang2024ayaexpanse} \citeyear{dang2024ayaexpanse}), Llama 4 Scout 17B Instruct (Llama; \citeauthor{meta2025llama4scout} \citeyear{meta2025llama4scout}), Mistral Magistral Small 24B (Magistral; \citeauthor{mistral2025magistral} \citeyear{mistral2025magistral}), Qwen3-Next-80B-A3B-Thinking (Qwen; \citeauthor{yang2025qwen3} \citeyear{yang2025qwen3}), DeepSeek-V3 (DeepSeek; \citeauthor{liu2024deepseekv3} \citeyear{liu2024deepseekv3}), and gpt-oss-20b (GPT-OSS; \citeauthor{openai2025gptoss20b} \citeyear{openai2025gptoss20b}). %All models follow the same protocol with decoding aligned for comparability. 
All models follow the experimental setup described in Section \ref{sec:prompts}.

\paragraph{Results} \label{sec:lang results}
Of the models tested, GPT-OSS is discarded because it is unable to produce consistent output. Aya and Llama are also discarded, as they return up to 10 and 9 labels respectively even when explicitly stating the number of labels in the prompt -- probably because of their limited reasoning capabilities compared to the other tested LLMs. Although Qwen and Magistral also sporadically return additional categories, we keep them in our experiments as the categories are justified by the model-generated explanations as a subset of the main ones (e.g., \textsc{personal identity} is mapped to \textsc{identity}).

\begin{table}
\centering
\small
\resizebox{1.0\linewidth}{!}{
\begin{tabular}{lcccccccc}
\toprule
 \textbf{Lang.} & \multicolumn{2}{c}{\textbf{DeepSeek}} & \multicolumn{2}{c}{\textbf{GPT-5}} & \multicolumn{2}{c}{\textbf{Magistral}} & \multicolumn{2}{c}{\textbf{Qwen}}\\
 &  F1w & F1m & F1w & F1m & F1w & F1m & F1w & F1m\\
\midrule
Luxemb. & \textbf{0.537} & \textbf{0.478} & 0.558 & 0.497 & 0.436 & 0.351 & 0.500 & 0.380 \\
German      & 0.518 & 0.469 & 0.551 & 0.500 & 0.450 & 0.381 & 0.529 & 0.399 \\
French      & 0.507  & 0.474 & 0.585 & 0.525 &  0.461  & \textbf{0.406} & \textbf{0.532} & \textbf{0.404} \\
English     & 0.490 & 0.446 & \textbf{0.596} & \textbf{0.542} & \textbf{0.462} & 0.346 & 0.514 & 0.392\\
\bottomrule
\end{tabular}}
\caption{\label{table comparison}Comparison of models per language (Weighted F1 and Macro F1)}
\end{table}

\autoref{table comparison} shows the performance of the remaining models broken down by language. While DeepSeek gets the highest scores with the original Luxembourgish data, the remaining models work slightly better with the translations. 
GPT-5 outperforms the other models in all languages, and the best overall performance is obtained with the English translations.

\subsection{Translation effects}

Zooming in on the two different sets of translations, we see that the quality of the translations does not affect model performance. For both translation sets, the best performing model is GPT-5 in all languages. As shown in \autoref{table automatic/post}, the GPT-5 performance gap between English and Luxembourgish is larger with automatic translations than with post-edited ones. For DeepSeek, Luxembourgish seems to work better than English both for automatic translations and post-edited ones.

\begin{table}[htbp]
%\centering
%\footnotesize
\resizebox{\linewidth}{!}{
\begin{tabular}{lcccccccc}
\toprule
 \multicolumn{1}{c}{\textbf{Lang.}} & \multicolumn{2}{c}{\textbf{DeepSeek}} & \multicolumn{2}{c}{\textbf{GPT-5}} & \multicolumn{2}{c}{\textbf{Magistral}} & \multicolumn{2}{c}{\textbf{Qwen}}\\
% & \multicolumn{2}{c}{F1w} & \multicolumn{2}{c}{F1w} & \multicolumn{2}{c}{F1w} & \multicolumn{2}{c}{F1w}\\
 & Auto & PE & Auto & PE & Auto & PE & Auto & PE \\ %& Auto & PE & Auto & PE & Auto & PE & Auto & PE \\
%\textbf{Language}\\
\midrule
Luxemb. & 0.762 & 0.771 & 0.797 & 0.780 & 0.699 & 0.737 & 0.755 & 0.661 \\
German   & 0.747 & \textbf{0.783} & \textbf{0.803} & 0.767 & 0.728 & \textbf{0.738} & \textbf{0.774} & 0.753\\
French  & 0.731 & \textbf{0.781} & \textbf{0.806} & 0.800 & 0.721 & \textbf{0.744} & \textbf{0.768} &  0.766 \\
English   & 0.743 & \textbf{0.750} & \textbf{0.829} & 0.783 & 0.724 & \textbf{0.746} & \textbf{0.767} & 0.740 \\
\bottomrule 
\end{tabular}
}
\caption{\label{table automatic/post} Comparison of models on Automatic (Auto) and Post-Edited (PE) translations across languages and on the original Luxembourgish corpus with the same split (Weighted F1)}
\end{table}

\begin{table*}[ht]
\centering
\resizebox{\linewidth}{!}{
\begin{tabular}{lcccccccccccc}
\toprule
\multicolumn{1}{c}{\textbf{Label}} & \multicolumn{3}{c}{\textbf{DeepSeek}} & \multicolumn{3}{c}{\textbf{GPT-5}} & \multicolumn{3}{c}{\textbf{Magistral}} & \multicolumn{3}{c}{\textbf{Qwen}} \\
& Precision & Recall & F1 & Precision & Recall & F1 & Precision & Recall & F1 & Precision & Recall & F1 \\
\midrule
Belonging      &0.468 & 0.603 & 0.527 & 0.541 & 0.492 & 0.516  & 0.466 & 0.329 & 0.386  & 0.365 & 0.468 & 0.410 \\
Identity       &0.338 & 0.526 & 0.411 & 0.504 & 0.362 & 0.421 & 0.216 & 0.612 & 0.319 & 0.341 & 0.459 & 0.391 \\
Recognition     & 0.502 & 0.218 & 0.304 & 0.479 & 0.608 & 0.536  & 0.447 & 0.092 & 0.153 & 0.458 & 0.348 & 0.395  \\
Responsibility   &0.603 & 0.242 & 0.345 & 0.689 & 0.211 & 0.323 & 0.667 & 0.190 & 0.296 & 0.368 & 0.295 & 0.328 \\
Vitality         & 0.554 &0.375 & 0.447 & 0.481 & 0.611 & 0.538 & 0.435 & 0.382 & 0.407 & 0.278 & 0.625 & 0.385\\
None         & 0.858 & 0.952 & 0.903 & 0.908 & 0.950 & 0.929 & 0.854 & 0.950 & 0.899 & 0.926 & 0.864 & 0.894\\ 
%macro avg &0.554 & 0.486 & 0.489 & 0.600 & 0.539 & 0.544 & 0.441 & 0.365 & 0.351 & 0.391 & 0.437 & 0.401\\
%weighted avg & 0.761 & 0.780 & 0.756 & 0.807 & 0.809 & 0.797 & 0.749 & 0.756 & 0.727 &0.771 & 0.736 & 0.748 \\
\bottomrule
\end{tabular}}
\caption{\label{table categories} Comparison of models per category}
\end{table*}
% \begin{table*}[ht]
% \centering
% \resizebox{\linewidth}{!}{
% \begin{tabular}{lcccccccccccc}
% \toprule
% \multicolumn{1}{c}{\textbf{Label}} & \multicolumn{3}{c}{\textbf{DeepSeek}} & \multicolumn{3}{c}{\textbf{GPT-5}} & \multicolumn{3}{c}{\textbf{Magistral}} & \multicolumn{3}{c}{\textbf{Qwen}} \\
% & Precision & Recall & F1 & Precision & Recall & F1 & Precision & Recall & F1 & Precision & Recall & F1 \\
% \midrule
% Belonging      &0.468 & 0.603 & 0.527 & 0.541 & 0.492 & 0.516  & 0.466 & 0.329 & 0.386  & 0.365 & 0.468 & 0.410 \\
% Identity       &0.338 & 0.526 & 0.411 & 0.504 & 0.362 & 0.421 & 0.216 & 0.612 & 0.319 & 0.341 & 0.459 & 0.391 \\
% Recognition     & 0.502 & 0.218 & 0.304 & 0.479 & 0.608 & 0.536  & 0.447 & 0.092 & 0.153 & 0.458 & 0.348 & 0.395  \\
% Responsibility   &0.603 & 0.242 & 0.345 & 0.689 & 0.211 & 0.323 & 0.667 & 0.190 & 0.296 & 0.368 & 0.295 & 0.328 \\
% Vitality         & 0.554 &0.375 & 0.447 & 0.481 & 0.611 & 0.538 & 0.435 & 0.382 & 0.407 & 0.278 & 0.625 & 0.385\\
% None         & 0.858 & 0.952 & 0.903 & 0.908 & 0.950 & 0.929 & 0.854 & 0.950 & 0.899 & 0.926 & 0.864 & 0.894\\ 
% %macro avg &0.554 & 0.486 & 0.489 & 0.600 & 0.539 & 0.544 & 0.441 & 0.365 & 0.351 & 0.391 & 0.437 & 0.401\\
% %weighted avg & 0.761 & 0.780 & 0.756 & 0.807 & 0.809 & 0.797 & 0.749 & 0.756 & 0.727 &0.771 & 0.736 & 0.748 \\
% \bottomrule
% \end{tabular}}
% \caption{\label{table categories} Comparison of Models per Category}
% \end{table*}
\paragraph{Discussion} 

To answer research question 2, translating to more-represented languages in NLP produces a limited performance improvement, whereas we find no improvement in the task between using automatic and post-edited translation sets. %a limited impact of translation and translation quality. %We consider two potential explanations for the limited impact of translation and translation quality:
We consider two potential explanations to this: cross-lingual transfer in LLMs and the challenge of transferring cultural and social meanings from the source text to the target languages. 

On the one hand, Luxembourgish shares some grammatical and lexical features with high-resource languages -- mainly German and French -- which are represented in the models' training data and officially supported by them. Therefore, it is quite likely that the tested models generalize from shared multilingual representations \cite{zhang-etal-2024-unveiling-linguistic}.

On the other hand, the way sociocultural meanings are treated in the translation process may also negatively impact the classification performance. For the post-edited set, we tried to find a middle ground between preserving the same language ideologies, reporting appropriate linguistic cues to each target language, and avoiding introducing noise in the translations. For instance, capitalization for emphasis is kept, while lack of orthographic capitalization is corrected in the translations. In the non-post-edited set, only mistranslated country and language mentions are corrected. However, our results do not show an increase in annotation performance that could be ascribed to the different translation conditions. Similarly, there is no significant performance improvement between the original Luxembourgish comments and the translations. This is attributable to the failure to eliminate noise and reproduce ideology in target languages. %Furthermore, transferring cultural elements of a specific culture into other languages could produce unexpected texts for the models. 

Due to these challenges, and to the LLMs' relatively even performances across languages, keeping the original text seems more appropriate.% in such settings.

\subsection{Label-specific results}

\autoref{table categories} shows F1 scores averaged on the four languages per category and model.
All analyzed models perform better for discriminating between items with and without language ideologies than for distinguishing between individual categories. In fact, the absence of language ideologies -- the label \textsc{none} -- is the category with higher F1 scores compared to the others (around 0.9). Compared to our gold standard, \textsc{responsibility} and \textsc{recognition} are under-detected, while \textsc{vitality}, \textsc{identity}, and \textsc{belonging} are over-predicted.

\begin{table}[ht]
\centering
\footnotesize
\begin{tabular}{lccc}
\toprule
\textbf{Label} & \textbf{DeepSeek} & \textbf{GPT-5} & \textbf{Highest} \\
\midrule
Belong.        & 0.487 & 0.512 & 0.571 (EN) \\
Identity         & 0.376 & 0.386 & 0.475 (FR) \\
Recogn.      & 0.330 & 0.520 & 0.565 (FR) \\
Respons.   & 0.398 & 0.299 & 0.398 (LU) \\
Vitality         & 0.448 & 0.500 & 0.555 (EN) \\
None         & 0.910 & 0.927 & 0.934 (EN) \\ 
\bottomrule
\end{tabular}
\caption{\label{table dp/gpt} Per-class F1-scores of DeepSeek and GPT-5.
The first two columns show F1-scores for DeepSeek and GPT-5 respectively, on the \textbf{Luxembourgish} dataset. The third column reports the absolute best F1 scores per label, regardless of the language of the dataset.}
\end{table}

% \begin{table}[ht]
% \resizebox{\linewidth}{!}{
% \begin{tabular}{lccc}
% \toprule
% \textbf{Label} & \textbf{DeepSeek} & \textbf{GPT-5} & \textbf{Highest} \\
% \midrule
% Belonging        & 0.487 & 0.512 & 0.571 (EN) \\
% Identity         & 0.376 & 0.386 & 0.475 (FR) \\
% Recognition      & 0.330 & 0.520 & 0.565 (FR) \\
% Responsibility   & 0.398 & 0.299 & 0.398 (LU) \\
% Vitality         & 0.448 & 0.500 & 0.555 (EN) \\
% None         & 0.910 & 0.927 & 0.934 (EN) \\ 
% \bottomrule
% \end{tabular}}
% \caption{\label{table dp/gpt} Per-class F1-scores of DeepSeek and GPT 5.
% The first two columns show F1-scores for DeepSeek and GPT5 respectively, on the Luxembourgish dataset. The third column reports the absolute best F1 scores per label, regardless of the language of the dataset.}
% \end{table}

DeepSeek and GPT-5 achieve the highest per-label F1 scores among the evaluated models. \autoref{table dp/gpt} shows per-label F1 scores for GPT-5 and DeepSeek on the original data, compared to the highest F1 score overall. GPT-5 has the highest scores in English for \textsc{belonging} (0.571) and \textsc{vitality} (0.555), and in French for \textsc{recognition} (0.565). DeepSeek has highest scores in French for \textsc{identity} (0.475) and in Luxembourgish for \textsc{responsibility} (0.398). These values are not significantly higher than the F1 scores obtained for Luxembourgish for DeepSeek and GPT-5, except for \textsc{recognition} for DeepSeek (0.330).

\paragraph{Analysis} Here, we examine the annotation categories in detail, focusing on the original Luxembourgish data and the Lang subset. 
We discuss consistent overlaps among categories, as well as potential causes of mislabeling. To understand the cause of systematic overlaps, we refer both to the content of the sentence and to the models-generated explanations. The benefit of models-generated explanations is discussed in the following section.
 
False negatives are observed for the category \textsc{recognition} in all other categories. Reading the explanations, the models seem to consider the hierarchy among languages -- which we subsume under \textsc{recognition} -- as a personal or national preference towards one or more languages. In the Luxembourg ideological landscape, the use of the pronoun \textit{our} (\textit{eis}) can indicate a reference to the national in-group. This information is provided as part of the \textsc{identity} category description. Therefore, also when not carrying an \textsc{identity} ideology, first-person plural pronouns and mentions of the ingroup are used as markers of national identity (Magistral and DeepSeek). Overlap with \textsc{vitality} arises when sentences describe the limited use of Luxembourgish in the country (Magistral, GPT-5, DeepSeek, Qwen). \textsc{recognition} is then mislabeled as (social) \textsc{belonging} when Luxembourgish is stated as the main language of the country (DeepSeek and Qwen). Furthermore, comments perceived as blaming the current language situation on actors mentioned in the text are labeled as \textsc{responsibility} (Qwen).\footnote{Further examples and statistics are provided in the appendix.}

For \textsc{responsibility}, it is mislabeled as \textsc{belonging} when the responsible actor is an out-group member (Qwen, GPT-5), and as \textsc{recognition} when it is an official institution or a private person (Qwen). It is then mislabeled as \textsc{vitality} when the implications of language decay are considered to be the main ideology of the item (Qwen, GPT-5, Magistral). Additionally, the use of first-person pronouns is reported in the model-generated explanations to justify the \textsc{identity} label (Magistral, DeepSeek). 

\textsc{Vitality} is mislabeled as \textsc{identity} when language decay is associated with national decay (DeepSeek), and through the use of first-person plural pronouns (Qwen). According to the model-generated explanations, items encouraging the use of the national language as a preservation mechanism are annotated as \textsc{recognition} (GPT-5).

\textsc{Belonging} is mislabeled as \textsc{identity} when Luxembourgish is stated as the language of society (Magistral, Qwen). The presence of the in-group in these utterances makes it difficult for the models to discriminate between the national language as a characteristic of the nation self-determination -- which we subsume under \textsc{identity} -- and as a means of social exclusion -- which we subsume under \textsc{belonging}.

Finally, \textsc{identity} is mislabeled as \textsc{recognition} when the model (GPT-5) fails to disambiguate between the use of language to create and reinforce a national feeling -- which we subsume under \textsc{identity} -- and the description of language use in the country -- which we subsume under \textsc{recognition}.

\begin{table*}
\centering
\resizebox{0.9\textwidth}{!}{
\renewcommand{\arraystretch}{1.3}
\begin{tabularx}{\textwidth}{|X|}
\hline
Dear -- We should use more Luxembourgish
words, and not German words, nor French words.\\ \hline
\textbf{Gold standard}: \textsc{recognition}\\ \hline
\textbf{Magistral} : \textsc{identity} \\'The sentence refers to the national group using the first-person plural pronoun ('mir') and emphasizes the use of Luxembourgish language in opposition to other languages (German and French). This aligns with the national identity category, as it highlights the importance of Luxembourgish language for Luxembourgers.' \\ \hline
\textbf{DeepSeek}: \textsc{vitality} \\'The sentence emphasizes the need to use Luxembourgish words instead of German or French words, which aligns with the 'vitality' category as it addresses the preservation of Luxembourgish language.'\\ \hline
\textbf{Qwen}: \textsc{vitality} \\'The sentence expresses concern about the use of German and French words instead of Luxembourgish words, which is a typical indicator of language decay and threat to the vitality of Luxembourgish. The advocacy for using more Luxembourgish vocabulary over foreign alternatives directly relates to preserving the language's integrity and resisting linguistic erosion.'\\ \hline
\textbf{GPT-5}: \textsc{vitality} \\'The speaker urges using more Luxembourgish words instead of German or French ones, proposing a way to safeguard the language against lexical replacement; this fits concerns and remedies typical of vitality.'\\ \hline

\end{tabularx}}
\caption{\label{tablemodel-gen-exp} Examples of models-generated explanations. The first line reports the annotated sentence, followed by labels and explanations by each model}%sentence ID 88, comment ID 3616 with assigned labels and generated explanations by the four evaluated models. Explanations are reported in inverted commas}
\end{table*}

\subsection{Examining model-generated explanations}

The language ideology definitions developed by the two annotators were given to the LLMs as part of the prompt. However, these guidelines were not always efficient, showing that defining language ideologies for LLMs introduces different challenges compared to human annotators.
For instance, since language ideologies might share both topics and linguistic features to express them, the annotators already faced (and solved) cases of overlapping categories. Despite providing specific guidelines, these cases are still error sources for LLMs -- together with unexpected ones. In this context, the explanations provided by the LLMs are particularly valuable, as they help us to understand how definitions might cause mislabeling, and how clearly delineated the categories are. 
%In this context, model-generated explanations are particularly valuable, as they provide insights into task setup improvements, helping to understand 1) how definitions might cause mislabeling, and 2) how clearly delineated the categories are.

\autoref{tablemodel-gen-exp} provides an example of a model-generated explanation. The annotated item is shown in its English translation, and is followed by the labels assigned by the evaluated models and the explanations. 
The \textsc{identity} label is justified by the use of the first-person plural pronoun \textit{we} (\textit{mir}) as a reference to the national group, and by the use of Luxembourgish in opposition to German and French. The \textsc{vitality} explanations show that inciting people to use more Luxembourgish words is considered a strategy against language decay. This example was annotated as \textsc{recognition} as it shows a desired hierarchy of language use opposed to the perceived status quo. However, ambiguity is induced by the exhortation of using Luxembourgish \textit{more}, that adds a hint to another ideological layer -- \textsc{vitality} -- to the language hierarchy. Similar examples caused label overlaps also for human annotators, but are already disentangled in the prompt given to the models. 

\paragraph{Language Ideology Definitions}
In general, clues which seem helpful for human annotators resulted in confusion between categories for the models. For instance, the use of specific pronouns and determiners is a useful hint for human annotators to discriminate between \textsc{identity} and other categories, but it causes an over-prediction of this category in the models -- as shown in the example above. Then, for \textsc{responsibility}, human annotators identified a subject of responsibility, as well as an object. Three different subjects of responsibility have been found -- in-group, politician, out-group -- and given to the model together with the description of the category. From the explanations, it seems that the presence of the three different responsibility agents leads the tested models to mislabel the category as \textsc{recognition} -- official institutions as subject of responsibility -- and \textsc{belonging} -- the out-group as subject of responsibility.\footnote{See \citet{gal_irvine_2019} for more information about language ideologies definitions.}

\paragraph{Language Ideology Labels} The analysis in the previous paragraphs suggests that the five language ideology labels might be too fine-grained for reliable prediction with simple in-context learning. Therefore, to answer research question 3: LLMs require expert support or task-specific model adaptation to adequately inform sociolinguistic research. This raises the question whether a more coarse-grained classification scheme would be beneficial. At the same time, the example also demonstrates how LLMs could serve as a valuable tool to improve human annotations and improve them, especially in cases of doubt.

For instance, both \textsc{identity} and \textsc{belonging}, although referring to two different social groups, convey feelings of social belonging obtained and constructed through language use. However, these ideologies highlight a power imbalance that will be lost if the categories are merged, i.e., the dominant group constructs its own identity -- \textsc{identity} -- and shapes the role of other people in society -- \textsc{belonging}. Similarly, statements about language preference could be grouped under one label,  but it will cause losing implications between \textsc{identity} -- personal preference -- and \textsc{recognition} -- reference to language status and hierarchy among languages. 
Therefore, restructuring the categorization scheme could mean losing important aspects of societal power dynamics for sociolinguistic analyses.

%On the other hand, using LLMs as tools to differentiate between utterances containing language ideologies and utterances without, already seems to be a quite useful task, mainly for selecting material to be analyzed. However, it might not help with further understanding language ideologies and language discourse in sociolinguistic research. 

%Additionally, models with lower scores already show in the explanations that they are not suitable for the task. For instance, Qwen and Magistral over-predict \textsc{identity} based on specific pronouns in the item more than DeepSeek and GPT-5. %or cut it!

\section{Conclusions}
This paper offers an evaluation of LLMs for language ideology detection. It shows that Luxembourgish does not need to be translated to higher-resourced languages in order to obtain satisfactory classification performance with generative LLMs.  
It highlights the crucial role of model-generated explanations for this task and that the evaluated models successfully discriminate items with language ideologies from those without. 
However, discriminating among the different categories of ideologies turns out to be more challenging. The first research question of the paper has thus both a positive and a negative answer: current generations of LLMs can perform a binary language ideology classification, but a fine-grained classification seems to be a more challenging task.
Although it might not help with further understanding language ideologies and language discourse in sociolinguistic research, a binary classification between content with and without language ideologies already seems to be a quite useful task, mainly for selecting material to be analyzed.
Our approach could be repurposed for other less-represented languages in NLP, to test if translations of original material could benefit the automatic analysis of language ideologies on the source language. However, given the complexity of the ideology classification task, this would quite likely result in an overview of the presence/absence of language ideologies in a discourse and, hence, require an in-depth analysis from trained (socio)linguists to understand how the motifs are implemented and whether additional language ideologies characterize the discourse in question. 

As for the nature of the task, language ideologies do not have a unique linguistic encoding: they may share topical vehicles, the context of utterance, as well as grammatical and lexical features, and are quite challenging for human annotators to discriminate. As including human annotators guidelines in the prompt design turned out to be of limited usefulness for LLMs, new paths should be considered. 
%As guidelines for human annotators turned out to be of limited usefulness for LLMs, new paths should be taken in consideration. 
One option, discussed above, could be the simplification of the annotation scheme. Another option could consist in collecting more manually annotated data to enable fine-tuning. The challenge will be to find a balance between sociolinguistic interests, LLM capabilities and human annotation efforts.

Although LLMs show promise as tools for sociolinguistic inquiry, their potential for the study of language ideologies remains limited without the guidance of human experts. Mainly, language ideologies cannot be studied only using LLMs with no previous understanding of the discourse in analysis -- as demonstrated in the trial phase. Mixed methods seem to be a promising way to enrich sociolinguistic research with quantitative methods and NLP advancements \cite{KircherHawkey2022, nguyen2016computational, purschke2025}.

In future experiments, we will focus on providing more general descriptions instead of detailed, example-based ones. We will improve them to highlight both core and specific elements per category, to give space to infrequent ideological expressions, and to reframe features frequently leading the models to mislabel certain categories. %Additionally, testing this approach with languages with similar status as Luxembourgish and structural similarities with higher-resourced language would help understand LLMs generalisation from big to small languages and their benefit for small languages. 

\section*{Limitations} 
As the project aimed to understand how two disciplines -- sociolinguistics and natural language processing -- can be mutually informative for each other, we are aware of limitations from both fields. As per the goal of the project, keyword searching should be adjusted to include language ideologies concerning Luxembourgish. Expanding the keyword search to include the different languages spoken in the country would help to get a more comprehensive view of language ideologies in the country.
Additionally, we refer to the LLM-generated annotations to understand what could be improved in the prompt design, but we are aware of their limitations in accurately describing the reasoning behind the classification \cite{randl2025mind, di2024explanation}.
Concerning the choice of models, two main limitations can be reported. First, models with limited reasoning capabilities proved to be not robust enough for this task and will be discarded for future experiments. Second, since we only had access to a small amount of annotated data, we were unable to train or fine-tune models on the task.
Future work should explore this direction, as improved model performance could support language discourse understanding, as well as its societal implications. However, although a fine-tuned model could achieve high performances, human annotators and linguistic experts remain essential to understand the consistency of model outputs. 

We noticed a grammatical mistake in the prompt used to preform this classification: for \textsc{identity} we state that it is expressed by the use of the first-person plural pronoun \textit{eis}. However, \textit{eis} is the dative form of the first-person plural pronoun \textit{mir}, and its possessive determiner. As this could be the cause of some labeling errors, we will implement the task further with the correct version of the prompt.

\section*{Ethical Statement} 
Data used for the project are provided and anonymized by the data owner RTL.lu.

\section*{Acknowledgments}
This research was supported by the Luxembourg National Research Fund (Project code: C22/SC/117225699). We would like to thank the members of the TRAVOLTA project and the Language and Technology Group of the University of Oslo for their invaluable advice on the project, and Raul Ian Sosa for his precious feedback. 

\section*{Bibliographical References}\label{sec:reference}

\bibliographystyle{lrec2026-natbib}
\bibliography{lrec2026-example}

% \section{Language Resource References}
% \label{lr:ref}
% \bibliographystylelanguageresource{lrec2026-natbib}
% \bibliographylanguageresource{languageresource}

\section*{Appendix A}
In this appendix, we report the original Luxembourgish version of the examples in Section \ref{sec:ideology_categories}:
\begin{itemize}
    \item \textbf{Example ID 636}: 1. Lëtzebuergesch ass fir eis Lëtzebuerger (250'000 Leit) eis éischt Sprooch. 
    \item \textbf{Example ID 3631}: Gudden Text, mee ech färten, eis Sprooch wärt ausstiewen.
    \item \textbf{Example ID 410165}: et ass geschafft:  All Awunner vu Lëtzebuerg hu sech sproochlech integréiert.
    \item \textbf{Example ID 666}: A wann eis Frontalieren kee Letzebuergesch schwaetzen dann ass daat eis eege schold.
    \item \textbf{Example ID 636}: 2. Lëtzebuergesch ass VIRUN ALLEM eng geschwaate Sprooch.
\end{itemize}

\section*{Appendix B}
Keywords used to retrieve the comments: \\ 'eis sprooch', 'eis sproch', 'ons sprooch', 'ons sproch', 'lëtzebuergesch', 'lëtzeboiesch', 'letzebuergesch', 'lëtzeburgesch', 'lëtzbuergesch', 'letzeburgesch', 'letzeboiesch', 'létzebuergesch', 'eiser sprooch', 'onser sprooch', 'eiser sproch', 'onser sproch', 'lëtzebuergescht', 'mammesprooch', 'mammesproch'.
\section*{Appendix C}
In this appendix, we add the complete version of Prompt 4:\\ \\
Detect one or more of the following categories in the sentences of the comment below, after the instructions and schemas.
The categories are: Identity, Vitality, Belonging, Responsibility, Recognition. These categories concern language ideologies and only these exact categories
can be used for the annotation.
No overlap between categories is allowed, each sentence can have 0 or 1 category.
If you cannot identify one of the category assign N/A.
Do not take any quoted parts (i.e. text in quotation marks) into account for the labeling.
For each sentence, explain why you have assigned that label.
Here are the explanations of the five categories: 

Identity: The language related ideology ‘identity’ groups both national and personal identity. 
National identity: it refers to Luxembourgish national identity and it only concerns Luxembourgish people, born in Luxembourg with Luxembourgish as native language. 
It is characterized by a strong emotional attachment to the language and culture of Luxembourg, in opposition to people with different languages and cultures. 
It is found in clauses or part of clauses underling:
\begin{itemize}
    \item the tied connection between national identity and national language (example: Luxembourgers unified by a common national language)
    \item the tied connection between Luxembourgish culture and Luxembourgish language (example: importance of Luxembourgish language and culture)
    \item the formation of the Luxembourgish national identity during the Second World War (Luxembourgers explicitly chose Luxembourgish language as national language for the first time)
    \item the opposition between the national group (= born and raised in Luxembourg, by family of Luxembourgish origins, with Luxembourgish culture and Luxembourgish as native language) and the group seen as external to the country (= not born in Luxembourgish families, or coming from abroad, or living across the borders but going to Luxembourg for work. These people are grouped by lack of knowledge of Luxembourgish language and refusal to learn it)
\end{itemize}
Grammatically, this category is expressed by the use of the first-person plural pronoun (‘eis’) when referring:
\begin{itemize}
    \item to the national group 
    \item to what belongs to the national group
    \item to what characterizes the national group.
\end{itemize}
‘eis’ is used in opposition to the third person plural pronoun (‘si’).
The third person plural pronoun refers to the group seen external to the country.
Personal identity = personal preference towards a language with no reference to any group-shared ideology or feeling (examples = speaking Luxembourgish as personal preference and not as a national identity trait) \\
examples: 
\begin{enumerate}
    \item lëtzebuergesch ass fir eis lëtzebuerger (250'000 leit) eis éischt sprooch.
    \item et gëtt leit de zu dachau agespaart ware wéint eiser sprooch an e vollek dat seng sproch verléiert verléiert och seng identitéit…
    \item ech perséinlech maan mer emmer en spaass draus vir mat de verschiddenen wierder ze spillen wann ech eppes schreiwen
\end{enumerate}
examples where this label is not applicable:
\begin{enumerate}
    \item tschüss!
    \item mee mir kann et dee moment egal sinn, well ech da wahrscheinlech an däitschland wunnen... tschüss!
    \item patricia courtois
\end{enumerate}

Vitality: The language related ideology ‘vitality’ groups opinions about Luxembourgish language considered:
\begin{itemize}
    \item endangered
    \item threatened
    \item in decay
    \item soon to be replaced by other languages
\end{itemize}
Threatening factors are:
\begin{itemize}
    \item languages other than Luxembourgish spoken in the country
    \item impossibility of speaking Luxembourgish in everyday contexts (shops, hospitals, …)
    \item languages other than Luxembourgish used at official levels
    \item lack of use or incorrect use of Luxembourgish orthographical and grammatical norms
    \item use of non-Luxembourgish words instead of available Luxembourgish words
    \item scarcity of Luxembourgish speaking people 
\end{itemize}
Proposing ways to preserve and safeguard Luxembourgish is also grouped in this category. The focal element of this category is always the language situation. Even when proposing possible ways to save the language and reporting the role policymakers have in this, the focus always stays on the language situation and does not shift on secondary elements.
There are a few semantic fields typical of this category: decay, endangerment, violence (towards the language). \\
examples:
\begin{enumerate}
    \item eng dagesmamm déi lëtzebuergesch schwätzt ze fannen, ass quasi onméiglech. awer eng babá fënnt ee séier
    \item soss gi mer all frankophon gemaach
    \item ech denken sou seier geht dat net verluer
\end{enumerate}
examples where this label is not applicable:
\begin{enumerate}
    \item bonjour a vill gléck a neie joer.
    \item all guddes an deem sënn.
    \item ech verstinn eent net,léif fra.
\end{enumerate}
Belonging:  The language related ideology ‘belonging’ groups answers to the following questions:
\begin{itemize}
    \item Do people born outside Luxembourg belong in Luxembourg? 
    \item Do people speaking Luxembourgish at a basic level belong to Luxembourg? 
    \item Do people not learning Luxembourgish belong to Luxembourg?
\end{itemize}
This category is about the integration in Luxembourgish society of people coming from abroad or people of non-Luxembourgish descents. It always involves the presence of foreigners, or of people not speaking Luxembourgish or not being Luxembourgish native speakers. 
Two main factors are significant in this category: 1) Luxembourgish language, 2) foreigners.
These two factors intertwine as follows:
\begin{itemize}
    \item Luxembourgish language is seen as the only means of integration in Luxembourgish society for foreigners
    \item Foreigners are expected to learn Luxembourgish 
    \item Foreigners refuse to learn Luxembourgish language 
    \item Luxembourgish language has to be learned at school for foreigners not to be disadvantaged in society
    \item The required level of Luxembourgish to obtain the citizenship is too low 
    \item Foreigners who learn or try to learn/speak Luxembourgish are welcomed in Luxembourgish society
    \item Foreigners adapting to Luxembourgish language and culture are part of Luxembourgish society
\end{itemize}
examples:
\begin{enumerate}
    \item eischt offiziell sprooch awer onbedengt fir e kloert zeeche fir d'integratioun ze setzen, fir eisen auslännesche matbierger kloer ze verstoen ze ginn dass mir eng eegestänneg denkweis, sprooch a kultur hunn an si sech eis unzepassen hunn andeems si net just 1 joer lëtzebuergesch léieren mee bis si et sou kënnen dass si et och kënne schwätzen a verstoen
    \item ween lëtzebuerger wëll ginn muss och lëtzebuergesch schwätzen
    \item ganz traureg daat do. wann et engem schlecht geht muss een sech kennen an senger mammesprooch ausdrecken an et huet een et nët néideg sech vun schlecht gelaunten franséischen infirmièren ungranzen ze loossen. déi däitsch maachen jo nach éischter en effort an probéieren lëtzebuergesch ze schwätzen
\end{enumerate}
examples where this label is not applicable:
\begin{enumerate}
    \item ierch all een schéinen owend
    \item där meenung sin ech och.
    \item wou soll daat dann hin feieren?
\end{enumerate}

Responsibility: The language related ideology ‘responsibility’ groups opinions about who is considered responsible for Luxembourgish decay. This category has two main elements: 1) responsible agent, 2) subject of responsibility.
The subject of responsibility is the Luxembourgish language decay (everything defined in ‘vitality’ category)
Possible responsible agents are:
\begin{itemize}
    \item Politicians and policymakers: because they don’t defend Luxembourgish language through adequate language policies and requires a low Luxembourgish language level to acquire Luxembourgish citizenship 
    \item Luxembourgers: because they never impose themselves to speak Luxembourgish with foreigners, they never ask foreigners to learn Luxembourgish, but adapt to foreigners with English, French, or German 
    \item Foreigners: because they refuse to learn/speak Luxembourgish and impose the languages they can speak (mainly French and English)
\end{itemize}
These responsible agents are also seen as responsible for a possible solution to the subject of responsibility.  
Other expressions fitting in this category involve Luxembourgers defending themselves from being called being racist, like ‘we are not racist, but…’, ‘they call us racist when we advocate for Luxembourgish.’ \\
examples:
\begin{enumerate}
    \item dann sollen dei ausländesch elteren emol ufänken eis sprooch ze leieren mee dofir sin mer ze faul
    \item et ass wéi am palais, do gët och keen lëtzebuergesch geschwaat.sie hun problemer vir sech an eiser sprooch ze artikuléieren
    \item et as leider net einfach ze verlaangen, dass soll letzebuergesch geschwaat soll gin, well soss get een nämlech ganz ganz ganz seier asl rassist ugesin. wann dir am geschäft engem soot e soll letzebuergesch schwetzen, da get een einfach lenks leihen gelos an färdeg
\end{enumerate}

examples where this label is not applicable:
\begin{enumerate}
    \item also muss daat och hei goen.
    \item schummt iech!
    \item as daat lo wierklech den sprengenden punkt vun desem artikel?
\end{enumerate}

Recognition: The language related ideology ‘recognition’ groups opinion on Luxembourgish language and other languages spoken in the country.
In this category, different ways of acknowledging languages are involved:
\begin{itemize}
    \item Luxembourgish considered as an oral language or/and a dialect
    \item Luxembourgish considered as an official, standardized language, whose norms have to be followed by speakers/writers
    \item Luxembourgish has to be recognized at the European level as one of the official languages of the European Union
    \item Hierarchy among languages 
\end{itemize}
examples: 
\begin{enumerate}
    \item mee ech verstinn haut nach ëmmer net, wéisou alles, awer och wierklech alles hei zu lëtzebuerg op franséisch leeft.
    \item jo mir hun eng nationalsprooch, an déi solle mir och héich halen, mee ons stäerkt war ëmmer, dass mir dräisproocheg waren, an och houfreg drop waren
    \item ass wuel net esou einfach, en dialekt forcement zu enger sprooch wëllen ze forméieren…
\end{enumerate}
examples where this label is not applicable:
\begin{enumerate}
    \item ganz richteg
    \item gudd geschriwen
    \item bai allem respekt awer dier sidd mengen ech am joer 1945 henken bliwen.
\end{enumerate}
examples where none of the labels described are applicable:
\begin{itemize}
    \item Ech kennen 2 lëtzebuergesch Mammen déi aleng mat hire Kanner do stin, déi kruten null Ennerstëtzung fir eng Wunneng ze fannen. 
    \item Letzebuerg huet keen mobiliteitsproblem ausser moies an owes zu den Spetzenstonnen.
    \item Kommt mir lossen eis net mei manipuleiren!
    \item Do huet de Fotograf awer gutt oppgepass fir emmer Leit opt Bild ze krein …
    \item gott sein dank dass mir keng vum "adr" an der spëtz hun soss hätten mir awer wirklech "FOLKLOR"
    \item Am EU parlament gouf gestemmt: 39 Jo- 478 neen-165 enthalung-72 abwesend nozekucken ob dem eu parlameter vun zdf. do kann een och kucken wien wei ofgestemmt hued.Letzebuergesch Parlamentarier hun all neen gestemmt ausser Mme Lulling.
\end{itemize}

%Return your response as a JSON object where each key is a sentence number from the sentences, the first value is the assigned label and the second label is the explanation.
%Here is a schema for your response. Continue counting up if there are more sentences.
%{
 %   "x": ["Label", "Explanation"],
  %  "y": ["Label", "Explanation"],
   % "z": ["Label", "Explanation"]
%}

%Here are the sentences of the comment in a JSON object, where each key is a sentence number and each value is a sentence. You must stick to the sentences and their numbers as they appear in the JSON object below.

\section*{Appendix D}
In this appendix, we add further examples of annotations mislabeled by all the models per category. The example is first reported in Luxembourgish followed by the English translation, its gold standard, and the mislabeled annotation by the models and model generated explanations. \\

\textbf{Example 586} \\ Mir sinn schon dat eenzegt Land wou d'Kanner mussen an enger anerer Sprooch alphabetiséiert ginn wéi hier Mammensproch, fir den net francophonen=Lëtzebuerger et lo nach méi schwéier ze machen, sollen d'Kanner lo och nach op franséisch alphabetiséiert ginn??\\
We are already the only country where the children have to be alphabetized in a language other than their mother tongue, to make it even more difficult for non-francophone=Luxembourgers, should the children be alphabetized in French??\\
\textbf{Gold standard}: \textsc{responsibility}\\ 
\textbf{Magistral} : \textsc{belonging}\\ 'The sentence discusses the integration of children from different linguistic backgrounds into Luxembourgish society, focusing on the role of Luxembourgish language in this process. It involves the presence of foreigners and the expectation to learn Luxembourgish.' \\
\textbf{DeepSeek}: \textsc{vitality} \\'The sentence emphasizes the need to use Luxembourgish words instead of German or French words, which aligns with the 'vitality' category as it addresses the preservation of Luxembourgish language.'\\ 
\textbf{Qwen}: \textsc{vitality} \\ 'The sentence expresses concern that children in Luxembourg are being alphabetized in a language other than their mother tongue (Luxembourgish), and further that non-French-speaking Luxembourgish children are being forced to learn in French, which is portrayed as making the situation even harder. This reflects a perception of Luxembourgish being threatened by the dominance of other languages in education — a core concern of the 'vitality' category, as it highlights endangerment through institutional language use.'\\ 
\textbf{GPT-5}: \textsc{recognition} \\ 'The sentence critiques alphabetizing children in a language different from their mother tongue and questions doing it in French, framing Luxembourgers as non-Francophone; this is an opinion about which languages should be used in official/educational contexts and the dominance of French.'

\textbf{Example 339} \\ Si sollen mol zu alleréischt all Lëtzebuergesch léieren. \\
First of all, they should all learn Luxembourgish. \\ 
\textbf{Gold standard}: \textsc{belonging}\\ 
\textbf{Magistral} : \textsc{identity}\\ 'The sentence emphasizes the importance of learning Luxembourgish first, which is a personal identity trait.' \\ 
\textbf{DeepSeek}: \textsc{responsibility} \\'The sentence emphasizes the responsibility of teaching Luxembourgish first, highlighting the importance of prioritizing Luxembourgish language education.'\\ 
\textbf{Qwen}: \textsc{identity} \\ 'The sentence emphasizes that children should learn Luxembourgish 'at first' ('zu alleréischt'), framing it as the foundational and primary language of the national group. The use of 'si sollen' (they should) implies a collective norm tied to national identity Luxembourgish as the essential, first language of Luxembourgers, reinforcing the tied connection between national identity and national language.'\\ 
\textbf{GPT-5}: \textsc{vitality} \\ 'Advocates that all children first learn Luxembourgish, emphasizing increased acquisition and priority of Luxembourgish as a way to strengthen and preserve the language.' 

\textbf{Example 356} \\ Jo mir hun eng Nationalsprooch, an déi solle mir och héich halen, mee ons Stäerkt war ëmmer, dass mir dräisproocheg waren, an och houfreg drop waren. \\
Yes, we have a national language, and we should hold it in high esteem, but our strength has always been that we were trilingual, and we were also proud of it. \\ 
\textbf{Gold standard}: \textsc{identity}\\ 
\textbf{Magistral} : \textsc{recognition}\\ 'The sentence acknowledges Luxembourgish as a national language and discusses the strength of being multilingual, which falls under the recognition category.' \\ 
\textbf{DeepSeek}: \textsc{recognition} \\'The sentence acknowledges Luxembourgish as the national language and emphasizes the importance of maintaining it, fitting the 'Recognition' category.'\\ 
\textbf{Qwen}: \textsc{recognition} \\ 'The sentence acknowledges Luxembourgish as a 'Nationalsprooch' that should be 'héich gehalen' (held high), while also recognizing the value of trilingualism ('dräisproocheg'). This reflects a hierarchical acknowledgment of languages — affirming Luxembourgish as official and worthy of prestige, while accepting coexistence with other languages. This fits the 'recognition' category, which deals with status, standardization, and hierarchy of languages.'\\ 
\textbf{GPT-5}: \textsc{recognition} \\ 'It explicitly discusses the status of Luxembourgish as a national language and values Luxembourg’s trilingualism, i.e., views on languages and their place in the country.' 

\textbf{Example 678} \\ Glaichzaiteg gehéiert Lëtzebuergesch zu onser Kultur a muss eis erhaale bleiwen!\\
GLikewise Luxembourgish belongs to our culture and must be preserved! \\ 
\textbf{Gold standard}: \textsc{vitality}\\ 
\textbf{Magistral} : \textsc{identity}\\ 'The sentence emphasizes the importance of Luxembourgish to the national culture and identity, which is characteristic of the 'Identity' category.' \\ 
\textbf{DeepSeek}: \textsc{identity} \\'The sentence emphasizes the importance of Luxembourgish to Luxembourg's culture, fitting the 'Identity' category as it highlights a connection between language and national identity.'\\ 
\textbf{Qwen}: \textsc{identity} \\ 'The sentence asserts that Luxembourgish belongs to 'our' culture and must be preserved, using the first-person plural 'eis' (our) to express collective emotional attachment to the language as a core element of national identity. This reflects the tied connection between national identity and national language, fitting the 'identity' category.'\\ 
\textbf{GPT-5}: \textsc{identity} \\ 'It explicitly links Luxembourgish to 'our culture' and uses first-person plural to assert a cultural-national attachment, a hallmark of the identity category.' \\ 

\section*{Appendix E}
In this appendix, we add row normalized confusion matrices for the Lang subset and for the whole dataset in analysis (Lang and NotLang) in its original language (Luxembourgish) for the four analyzed models: GPT-5, DeepSeek, Magistral, and Qwen.
%\begin{figure}[h]
%    \centering
%    \includegraphics[width=0.9\linewidth]{confmatrix_GPT5_rownormalised.png}
%    \caption{Confusion matrix per annotation with GPT-5 - Lang subset}
%    \label{fig:GPT5_rownorm}
%\end{figure}
%\begin{figure}
%    \centering
%    \includegraphics[width=0.9\linewidth]{confmatrix_GPT5_complete.png}
%    \caption{Confusion matrix per annotation with GPT-5 - Lang and Notlang subsets}
%    \label{fig:GPT5_complete}
%\end{figure}
%\begin{figure}[h]
%    \centering
%    \includegraphics[width=7cm]{confmatrix_GPT5.png}
%    \caption{Confusion matrix per annotation with GPT-5 - globally normalised Lang subset}
%    \label{fig:gpt}
%\end{figure}

\begin{figure} [h]

\begin{subfigure}{0.5\textwidth}
\centering
\includegraphics[width=\linewidth, height=6cm]{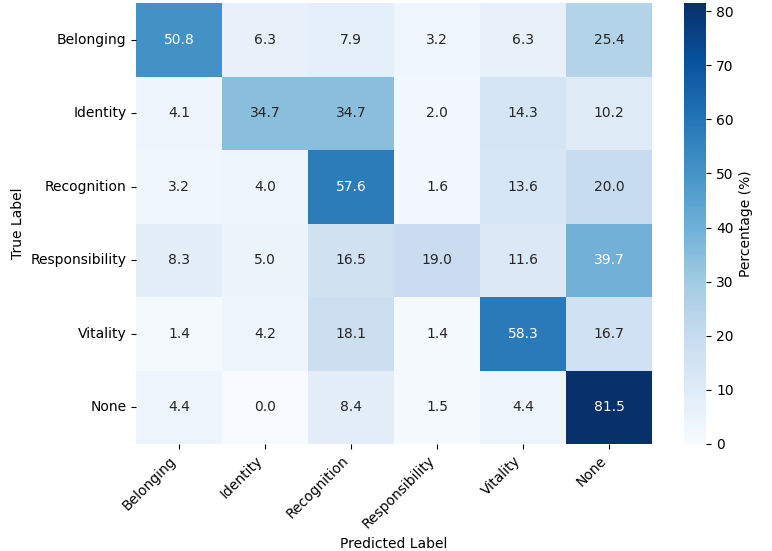} 
\caption{Lang subset}
\label{fig:GPT5_rownorm}
\end{subfigure}
\hfill
\begin{subfigure}{0.5\textwidth}
\centering
\includegraphics[width=\linewidth, height=6cm]{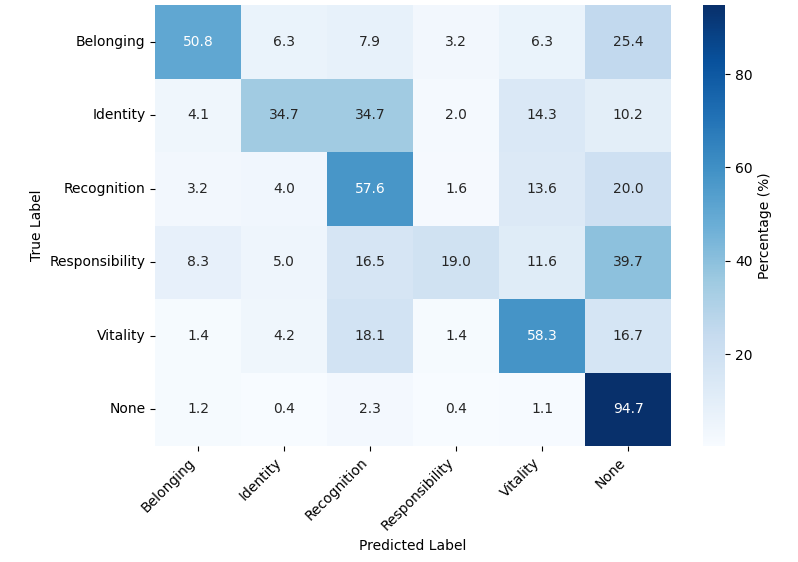}
\caption{Lang and NotLang subsets}
\label{fig:GPT5_complete}
\end{subfigure}

\caption{Confusion matrix per annotation with GPT-5}
\label{fig:image2}
\end{figure}

%\begin{figure}
%    \centering
%    \includegraphics[width=0.9\linewidth]{confmatrix_DEEPSEEK_rownormalised.png}
%    \caption{Confusion matrix per annotation with DeepSeek - Lang subset}
%    \label{fig:DeepSeek_rownorm}
%\end{figure}
%\begin{figure}
%    \centering
%    \includegraphics[width=0.9\linewidth]{confmatrix_DEEPSEEK_complete.png}
%    \caption{Confusion matrix per annotation with DeepSeek - Lang and NotLang subsets}
%    \label{fig:DeepSeek_complete}
%\end{figure}
%\begin{figure}[h]
%    \centering
%    \includegraphics[width=7cm]{confmatrix_DEEPSEEK.png}
%    \caption{Confusion matrix per annotation with DeepSeek - globally normalised}
%    \label{fig:deepseek}
%\end{figure}

\begin{figure}

\begin{subfigure}[b]{0.5\textwidth}
\centering
\includegraphics[width=\linewidth, height=6cm]{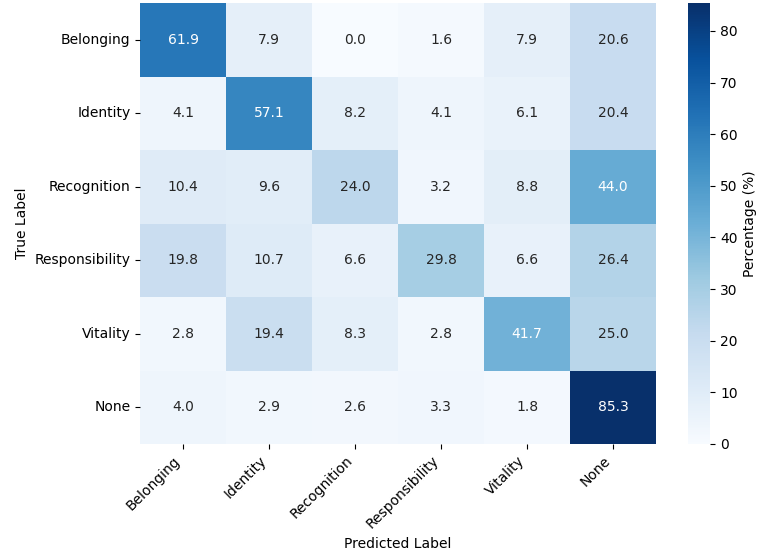} 
\caption{Lang subset}
\label{fig:DeepSeek_rownorm}
\end{subfigure}
\hfill
\begin{subfigure}[b]{0.5\textwidth}
\centering
\includegraphics[width=\linewidth, height=6cm]{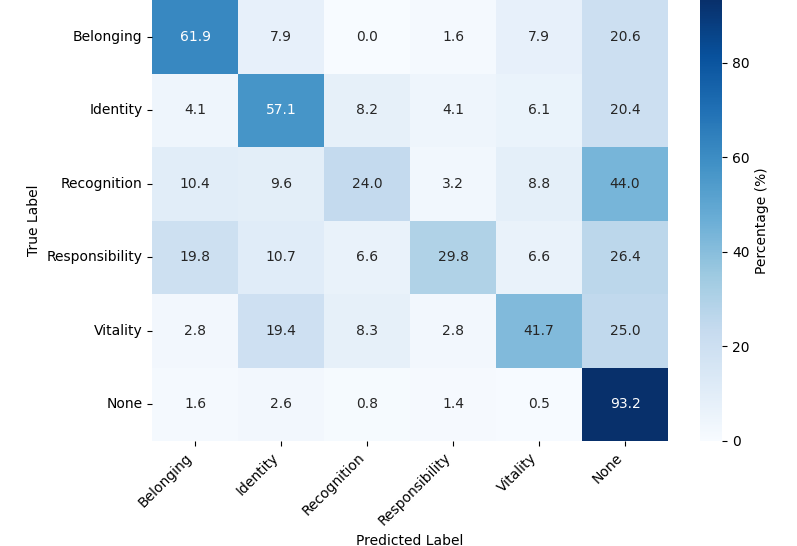}
\caption{Lang and Notlang subsets}
\label{fig:DeepSeek_complete}
\end{subfigure}

\caption{Confusion matrix per annotation with DeepSeek}
\label{fig:image3}
\end{figure}

%\begin{figure}
%    \centering
%    \includegraphics[width=0.9\linewidth]{confmatrix_MAGISTRAL_rownormalised.png}
%    \caption{Confusion matrix per annotation with Magistral - Lang subset}
%    \label{fig:magistral_rownorm}
%\end{figure}
%\begin{figure}
%    \centering
%    \includegraphics[width=0.9\linewidth]{confmatrix_MAGISTRAL_complete.png}
%    \caption{Confusion matrix per annotation with Magistral - Lang and NotLang subsets}
%    \label{fig:magistral_complete}
%\end{figure}
%\begin{figure}[h]
%    \centering
%    \includegraphics[width=7cm]{confmatrix_MAGISTRAL.png}
%    \caption{Confusion matrix per annotation with Magistral - globally normalised - Lang subset}
%    \label{fig:magistral}
%\end{figure}

\begin{figure}[h]

\begin{subfigure}[b]{0.5\textwidth}
\centering
\includegraphics[width=\linewidth, height=6cm]{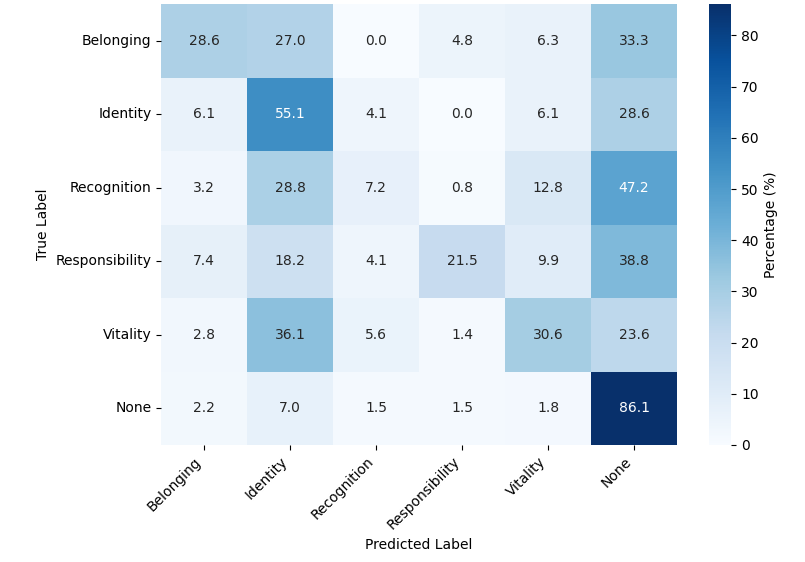} 
\caption{Lang subset}
\label{fig:magistral_rownorm}
\end{subfigure}
\hfill
\begin{subfigure}[b]{0.5\textwidth}
\centering
\includegraphics[width=\linewidth, height=6cm]{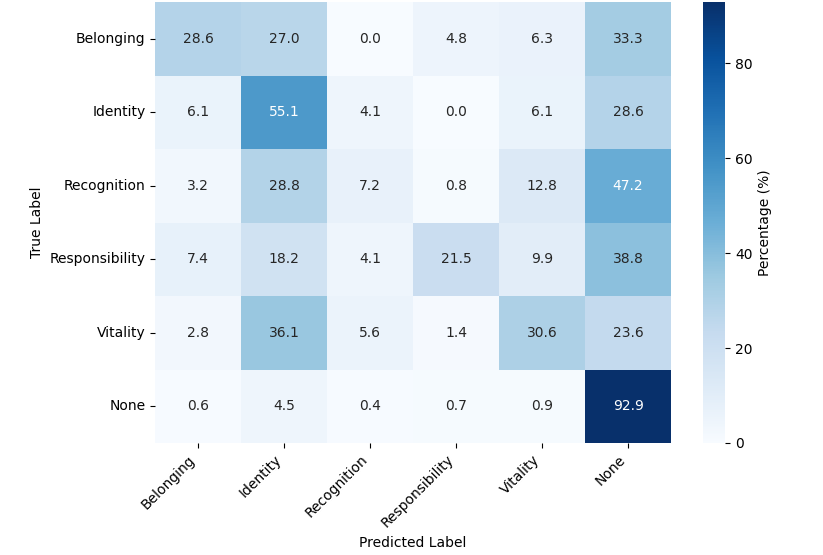}
\caption{Lang and NotLang subsets}
\label{fig:magistral_complete}
\end{subfigure}

\caption{Confusion matrix per annotation with Magistral}
\label{fig:image4}
\end{figure}

%\begin{figure}
%    \centering
%    \includegraphics[width=0.9\linewidth]{confmatrix_QWEN_rownormalised.png}
%    \caption{Confusion matrix per annotation with Qwen - Lang subset}
%    \label{fig:qwen_rownorm}
%\end{figure}
%\begin{figure}[ht]
%    \centering
%    \includegraphics[width=0.9\linewidth]{confmatrix_QWEN_complete.png}
%    \caption{Confusion matrix per annotation with Qwen - Lang and NotLang subsets}
%    \label{fig:qwen_complete}
%\end{figure}
%\begin{figure}[t]
%    \centering
%    \includegraphics[width=7cm]{confmatrix_QWEN.png}
%    \caption{Confusion matrix per annotation with Qwen - globally normalised - Lang subset}
%    \label{fig:qwen}
%\end{figure}

\begin{figure}[h]

\begin{subfigure}{0.5\textwidth}
\includegraphics[width=\linewidth, height=6cm]{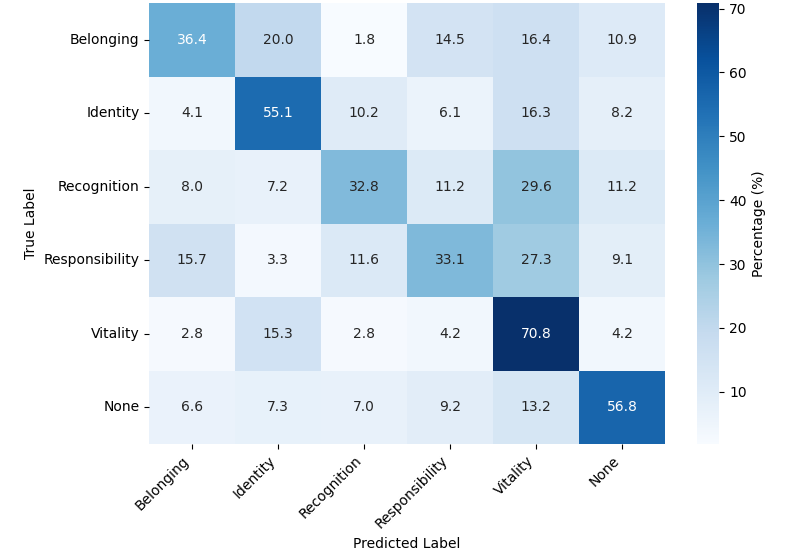} 
\caption{Lang subset}
\label{fig:qwen_rownorm}
\end{subfigure}
\begin{subfigure}{0.5\textwidth}
\includegraphics[width=\linewidth, height=6cm]{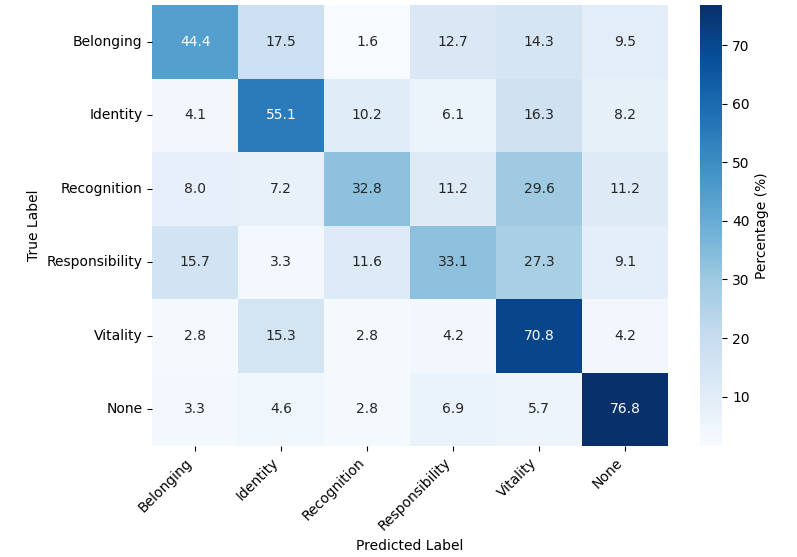}
\caption{Lang and NotLang subsets}
\label{fig:qwen_complete}
\end{subfigure}

\caption{Confusion matrix per annotation with Qwen}
\label{fig:image5}
\end{figure}

\end{document}